\documentclass[10pt,twocolumn,letterpaper]{article}

\usepackage{cvpr}
\usepackage{times}
\usepackage{epsfig}
\usepackage{graphicx}
\usepackage{amsmath}
\usepackage{amssymb}

\usepackage{caption}
\usepackage{subcaption}

\usepackage{makeidx}
\usepackage{color}
\usepackage{listings}
\usepackage{array}
\usepackage{dsfont}
\usepackage{multirow}
\definecolor{gray}{rgb}{0.5,0.5,0.5}

\usepackage{algpseudocode}
\usepackage{algorithm}
\usepackage{xspace}
\graphicspath{{./images/}}

\newcommand{\myparagraph}[1]{\vspace{6pt}\noindent{\bf #1}}

\def\1{\mathds{1}}

\makeatletter
\DeclareRobustCommand\onedot{\futurelet\@let@token\@onedot}
\def\@onedot{\ifx\@let@token.\else.\null\fi\xspace}

\def\eg{{e.g}\onedot} 
\def\ie{{i.e}\onedot} 
 
\def\etc{{etc}\onedot}

\g@addto@macro\normalsize{
  \setlength\abovedisplayskip{7pt}
  \setlength\belowdisplayskip{7pt}
  \setlength\abovedisplayshortskip{5pt}
  \setlength\belowdisplayshortskip{5pt}
}

\makeatother

\newcommand{\mateusz}[1]{#1}

\usepackage{framed}
\usepackage{xcolor}
\definecolor{gainsboro}{RGB}{220,220,220}

\usepackage[pagebackref=true,breaklinks=true,letterpaper=true,colorlinks,bookmarks=false]{hyperref}

\cvprfinalcopy 

\def\cvprPaperID{1127} 

\ifcvprfinal\fi
\begin{document}

\title{Multi-Cue Zero-Shot Learning with Strong Supervision}

\author{Zeynep Akata, Mateusz Malinowski, Mario Fritz and Bernt Schiele \vspace{4mm} \\ 
{Max-Planck Institute for Informatics}
}

\maketitle
\begin{abstract}
Scaling up visual category recognition to large numbers of classes remains challenging. A promising research direction is zero-shot learning, which does not require any training data to recognize new classes, but rather relies on some form of auxiliary information describing the new classes. Ultimately, this may allow to use textbook knowledge that humans employ to learn about new classes by transferring knowledge from classes they know well. The most successful zero-shot learning approaches currently require a particular type of auxiliary information -- namely attribute annotations performed by humans -- that is not readily available for most classes. Our goal is to circumvent this bottleneck by substituting such annotations by extracting multiple pieces of information from multiple unstructured text sources readily available on the web. To compensate for the weaker form of auxiliary information, we incorporate stronger supervision in the form of semantic part annotations on the classes from which we transfer knowledge. We achieve our goal by a joint embedding framework that maps multiple text parts as well as multiple semantic parts into a common space.
Our results consistently and significantly improve on the state-of-the-art in zero-short recognition and retrieval.
\end{abstract}


\begin{figure}[t]
\begin{center}
\includegraphics[width=0.9\linewidth, trim=10 30 20 0]{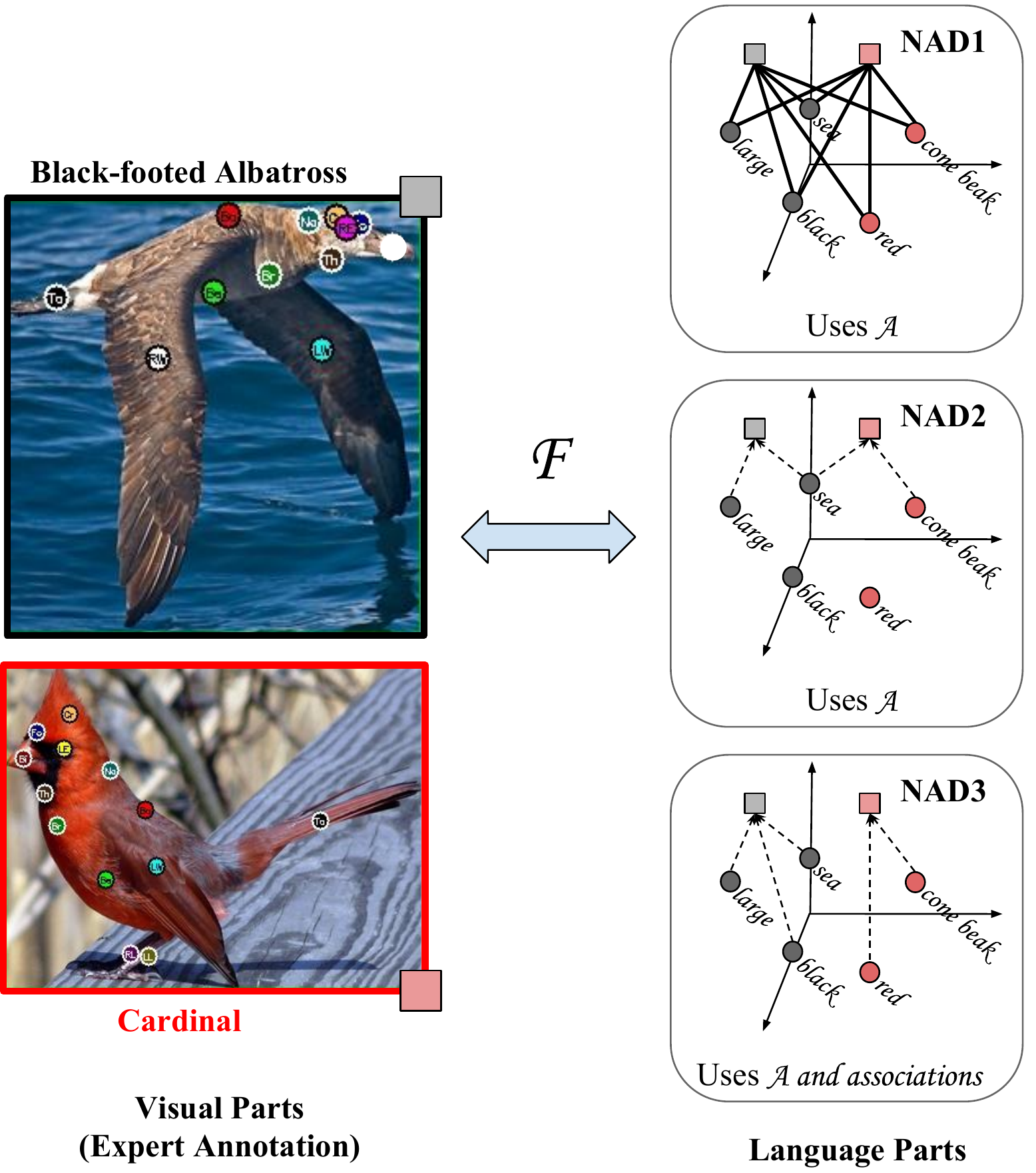}
\end{center}
\caption{We propose to jointly embed multiple language representations and multiple semantic visual parts for fine-grained zero-shot recognition. Our novel Noun-Attribute-Difference (NAD) representations are based on differences and distances between vectors in the word2vec space.}
\label{fig:DeFrag}
\end{figure}

\section{Introduction}
\label{sec:introduction}
The acquisition of visual concepts in humans and machines is still very different. It is hypothesized that early concept acquisition in children mostly follows a learning by example approach where visual concepts are directly grounded in sensory information and linked across modalities~\cite{roy2002learning}. However, this alone does not explain the diverse visual knowledge of an adult. A lot of our knowledge is preserved and conveyed via text and nowadays online resources. This enables humans to recognize objects without ever having seen a single instance of that object. Therefore the knowledge of the class is no longer solely grounded in sensory information, but rather transferred from prior experiences to new classes. A practical example includes field guides that describe different animal species via a range of categorizations and part descriptions~\cite{CaltechUCSDBirdsDataset} and allow to recognize an animal without expert knowledge.

Recent work on zero-shot learning for visual recognition aims at equipping computer vision systems to recognize novel classes without a single training example. The required ``knowledge'' for the recognition task is transferred via auxiliary information of different types. The most successful techniques utilize human annotations of attributes for each class. This is a particular type of auxiliary information that is not readily available in large quantities. This limitation hampers the progress of large-scale zero-shot learning. We argue that ultimately zero-shot learning techniques should leverage the same ``auxiliary information'' in terms of text books and online articles that humans use, as those are readily available in large quantities.

Realizing that such sources might be more noisy and difficult to leverage, we propose to combine them and to additionally use existing strong supervision for the visual information. In particular, on the fine grained recognition task that we investigate, detailed semantic part annotations are available and on other datasets such as Pascal 3D~\cite{xiang14wacv} detailed annotations of key-points are available.
Such information provides strong visual supervision and has shown to greatly improve recognition accuracy \cite{ZDGD14,oh2015person}. Our goal is to compensate the loss in performance by using weaker -- but more broadly available -- auxiliary language information with a stronger visual supervision for classes that we are transferring from.

Following the multi-modal embeddings paradigm for zero-shot learning~\cite{WBU11,FCSBM13,APHS15,ARWLS15,BSFS15}, we build a new framework that uses strong visual supervision in an embedding formulation that is flexible enough to accommodate a wide range of textual sources.
Our contributions are as follows.
(1) We propose to adapt Deep Fragment Embeddings~\cite{KL15} used for language generation to zero-shot learning facilitating a joint embedding of multiple language cues and visual information into a joint space. Our framework supports and integrates a wide range of textual and visual sources.
(2) We propose a novel language embedding method leveraging unstructured text as well as attributes without requiring any human annotation.
(3) We use strong supervision in terms of semantic part annotations to compensate for weaker but more broadly available auxiliary language information. We improve the state-of-the-art for fine-grained zero-shot learning, both using unsupervised text sources as auxiliary information and supervised attribute annotations if available.
(4) We show that the use of stronger visual annotations during training allows to improve zero-shot performance
without requiring the same strong supervision during recognition.

The rest of the paper is organized as follows. Sec~\ref{sec:related_work} summarizes related work, Sec~\ref{sec:deFrag} details our multi-modal embedding framework for zero-shot learning with strong supervision, Sec~\ref{sec:visual} presents our motivations for using visual parts as strong supervision, Sec~\ref{sec:language} details existing and proposed text embedding methods for language parts, Sec~\ref{sec:experiments} presents our experiments, and Sec~\ref{sec:conclusion} concludes.

\section{Related Work}
\label{sec:related_work}

Our work handles the challenging zero-shot problem~\cite{YA10,RSS11,KKTH12,LNH13,APHS15,NMBSSFCD13,FHXFG14} of the lack of labeled training data. Since the training and test classes are disjoint, traditional supervised learning methods which require per-image class labels cannot be directly applied. Therefore, side information that models the relationship between classes is required.

Attributes~\cite{FZ07,FEH10,LNH13} relate different classes through well-known, shared and human-interpretable traits. They are often collected manually~\cite{KKTH12,PG11,DPCG12} and have shown promising results for image classification~\cite{LNH13,APHS15}. On the other hand, the attribute collection through human annotations becomes a costly process for fine-grained data collections~\cite{CaltechUCSDBirdsDataset} where often only subtle visual differences between the objects exist. Therefore one needs a large number of attributes some of which can only be recognized and discriminated by field experts. This greatly increases the cost of annotations. Side information can also be collected automatically~\cite{MSCCD13,H54} from a large text corpora such as wikipedia. Word2vec~\cite{MSCCD13} learns a word's representation based on the word occurrence statistics, BoW~\cite{H54} uses a pre-defined vocabulary to build word histograms. Label embedding methods~\cite{BWG10} have been shown effective in modeling latent relationships between classes. For optimizing a multi-class classification objective through label embeddings, WSABIE~\cite{WBU10} uses images and corresponding labels to learn a label embedding. For zero-shot learning, DeViSE~\cite{FCSBM13} employs a ranking based bi-linear label embedding objective with image and distributed text representations as input/output embeddings. Similarly, ALE~\cite{APHS15} employs an approximate ranking objective that uses images and class-based attributes. ConSe~\cite{NMBSSFCD13} uses the probabilities of a softmax-output layer to weigh the semantic vectors of all the classes. \cite{ARWLS15} evaluates class-based vector representations built on fine-grained datasets for the zero-shot setting.

Similar embedding principles, often combined with recurrent neural networks~\cite{hochreiter97nc} or a dependency parser~\cite{dependency_parser}, have recently been applied to image-to-text retrieval~\cite{mao2014explain,KJL14,klein2015fisher}, language generation~\cite{VTBE15,KL15,DHGRVSD15}, and question answering about images~\cite{malinowski2015ask,gao2015you,ren2015image}.
Our work follows the latest research in joint modeling of language and vision features by formulating an  embedding of visual and textual representations in a joint space. In contrast to prior work, our approach accommodates and effectively integrates a wide range of textual representations and uses strong supervision in the form of semantic parts that remain optional at test time. In other words, we combine the advantages of two frameworks, \ie joint image-text embeddings for zero-shot learning~\cite{APHS15,ARWLS15} and sentence generation through pairwise similarity between visual and textual fragments~\cite{KJL14,KL15}, within a unified framework.

\section{Zero-Shot Multi-Cue Embeddings}
\label{sec:deFrag}

Following the state of the art zero-shot classification approaches in visual recognition \cite{WBU11,FCSBM13,APHS15,ARWLS15,BSFS15}, we cast image classification as learning a compatibility function between images and their textual descriptions.
The best known results have been obtained using human attribute descriptions \cite{ARWLS15}, which limits the applicability of zero shot approaches due to the necessity of human intervention. Consequently, there is a desire to replace such human input and transition to an unsupervised setting that only leverages data readily available, e.g. from online text sources.
Yet, prior evaluation~\cite{ARWLS15} of such unsupervised approaches, which  use data automatically extracted from large text corpora, have shown a significant drop in accuracy. While resources like wikipedia most likely contain more information on a target class than a few human-annotated attributes, it has not yet been possible to leverage them to their fullest.

To better leverage readily available textual sources in an unsupervised setting, we argue for holistic embedding techniques that combine multiple and diverse language representations together in order to capture the content of rich textual sources in multiple textual parts  allowing for a better transfer of knowledge to unknown classes. 
Additionally, we suggest a stronger supervision on the visual side, e.g. in terms of semantic part annotations, that extracts visual information from the known classes.
In the following, we will present our embedding formulation that achieves both objectives.

We map semantic visual parts and language parts into a common embedding space by combining the compatibility learning framework based on embeddings~\cite{FCSBM13,APHS15,ARWLS15} and the Deep Fragment embeddings (DeFrag~\cite{KL15}) objective in a single framework for zero-shot learning.

\paragraph{Objective.} We define a zero-shot prediction function that for given visual input ($x$), chooses the corresponding class ($y$) with the maximum compatibility score:
\begin{align} 
f(x) = \arg\max_{y} F(x,y).
\end{align} 
The compatibility function $F$ is defined over the language and visual parts as follows:
\begin{align}
F(x,y) = \frac{1}{|g_x||g_y|} \sum_{i\in g_x}\sum_{j\in g_y} \max(0,v_i^Ts_j)
\end{align}
where $g_x$ is \mateusz{a} set of visual parts for \mateusz{the} image $x$ and $g_y$ is \mateusz{a} set of language parts describing class $y$.
We define our multi-cue language and visual part embeddings as follows: 
\begin{align}
& s_j = f\left( \sum_m W^{\text{language}}_m l_m + b^{\text{language}} \right) \nonumber \\
& v_i = W^\text{visual} [CNN_{\theta_c}(I_b)] + b^{\text{visual}}
\end{align}
$l_m$ is a token from a language modality $m$ (we use human annotated class-attributes, word2vec, and BoW as language cues in our experiments), and all $W^{\text{language}}_m$ are the encoders for each modality that embed the language information into a joint space. $f(.)$ is the Rectified Linear Unit (ReLU) which computes $x \leftarrow \max (0,x)$. $CNN(I_b)$ denotes a part descriptor extracted from the bounding box $I_b$ surrounding the image part annotation $b$ using deep convolutional neural networks. The extracted descriptor is subsequently embedded into the space of visual parts via the encoder $W^\text{visual}$. The $\max$ is truncated at $0$ because the scores that are greater than $0$ are considered as correct assignments.

Finally, our objective function takes the form:
\begin{equation}\label{eq:obj}
C(\theta) = C_P(\theta) + \alpha \|\theta\|^2_2
\end{equation}
with $\theta = \{W^{language}, W^{visual}\}$ being the parameters of the framework and the constraints are defined as:
\begin{equation}\label{eq:ranking_constraints}
F(x_n, y) + \Delta \leq F(x_n, y_n), \forall{y \in \mathcal{Y}} 
\end{equation}
where $(x_n,y_n)$ denotes corresponding image-class pairs available during training. Intuitively, we optimize for a compatibility function that scores higher by at least a margin of $\Delta$ for true image-class pairs.

The part alignment objective ($C_P$) in Eq~\ref{eq:obj} enforces a language part to have a high score if that language part is relevant to the image: 
\begin{align}\label{eq:frag_align}
C_P(\theta) & = \sum_i\sum_j max(0,1-y_{ij}v_i^Ts_j)
\end{align}
In practice, we solve Eq. \ref{eq:frag_align} via $y_{ij} := sign(v_i^T s_j)$, a heuristic for Multiple Instance Learning~\cite{andrews2002support} as it offers an efficient alternative to direct optimization.

\myparagraph{Optimization.} The objective function (Eq.~\ref{eq:obj}) is optimized with Stochastic Gradient Descent (SGD) with mini-batches of 100, momentum 0.9, 
and 20 epochs through the data. We learn the word vectors $l_m$ and part descriptors $CNN(I_b)$ once and keep them fixed during the entire optimization procedure. We validate the margin $\Delta$, the learning rate and the dimensionality of the embedding space based on the accuracy on a validation set.

\section{Semantic Visual Parts}
\label{sec:visual}

Using parts for visual recognition has a long and successful history for general 
object recognition including~\cite{FergusIJCV2007,leibe08ijv,FGAR10}. 
The notion of semantic parts plays a central role in domains such as 
human pose estimation~\cite{yang13pami,chen14nips}, action recognition~\cite{DesaiECCV2012,leonid14gcpr,Cheron2015} and face detection~\cite{ZhuCVPR2012}.
For fine-grained classification~\cite{WF09,LNH13,KBBN09,WM10} where several object parts are shared across categories, discriminative parts are important for good performance~\cite{ZDGD14,ZPRDB14}. 
Also, CNNs have been shown to implicitly~\cite{ZKLOT15,ZLAXTO14} model discriminative parts of objects and images.

Based on the success of using parts in various forms for object recognition 
we hypothesize that using strong supervision in the form of (semantic) part annotations
should help fine-grained zero-shot learning. Also intuitively, an object class 
can be determined given the visual parts that it is composed of, \eg a \textit{large} sea bird with \textit{black feet} and \textit{curved beak} is a \textit{black footed albatross}.
Therefore, using strong supervision in the form of part annotations we seek to mitigate the loss of accuracy by using weaker auxiliary information in the form of unsupervised language representations. 
Note, that in this work we only rely on having the part positions annotated but do not use any other information such as part name or part type. In fact, the objective of our embedding method is formulated so that it does not require such one-to-one correspondence between textual and visual parts. 

More specifically, in this work, we use a pre-trained deep convolutional network (CNN) to extract multiple semantic visual parts from 19 bounding boxes surrounding different image part annotations, \ie the whole image, head, body, full object, and 15 part locations annotated by fine-grained object experts~\cite{CaltechUCSDBirdsDataset}.  

\section{Language Parts}
\label{sec:language}
Zero-shot learning approaches have been struggling to carry over the success from human attribute annotations to less explicit but readily available descriptions like wikipedia articles.
In order to advance the transfer of class knowledge, we study a wide range of language part representations that all can be accommodated by our embedding approach.
We investigate traditional human attribute annotations, two established word vector extraction methods, word2vec and BoW, as well as propose two novel methods as an improvement of these two, NAD and MBoW.

\subsection{Prior Representations of Attributes and Text}
\label{subsec:baseline}

Attributes~\cite{LNH13} are distinguishing properties of objects, \eg curved beak, eats planktons, lives in water, \etc that are easily recognized by humans and interpreted by computers.
Attributes are typically obtained through a two-step manual annotation process.
The first step is building a set of distinguishing properties that are related to a specific class while the second step is about rating the presence/absence or the strength of every attribute for each class.
In the context of fine-grained data collections, as most of the properties are common across categories, the number of distinguishing visual properties required is large which increases the annotation cost.
We refer to them in our experiment as the supervised scenario and aim to develop solutions for the unsupervised scenario where such human annotations are not necessary anymore.

Word2vec~\cite{MSCCD13} maps frequently occurring words in a document to a vector space.
It is a two-layer neural network that learns to predict a set of target words from a set of context words within a context window.
Word2vec summarizes a document and converts it into a vector. 
In our case, one class, \eg black footed albatross, is one document and therefore can be represented as a vector.
Word2vec has been previously shown~\cite{ARWLS15} to be effective for image classification and even fine-grained visual recognition.
We use existing~\cite{ARWLS15} fine-grained class-word2vec vectors for direct comparison of their and our frameworks.
Bag-of-Words is constructed as a per-class histogram of frequently occurring words. We use wikipedia documents that corresponds to the class of interest.
The vocabulary of frequently occurring words is defined by counting the number of frequently repeating words inside the entire document that contains all the classes.
The least and most frequently occurring words are eliminated from the vocabulary due to their irrelevance or redundancy.
We use the BoW vectors of~\cite{ARWLS15} for a fair comparison.

\subsection{NAD: Noun-Attribute-Differences} 
\label{subsec:multi-lang_w2v}
Parallel to our multiple visual parts argument, we aim to exploit semantic relationships of different words
to derive multiple language parts.
Word2vec (Sec~\ref{subsec:baseline}) builds a vectorial representation of each word that belongs to a learned vocabulary.
The word2vec vector space is constructed with the aim to capture semantics and 
as a result
word2vec captures several semantic regularities~\cite{MCCD13,LG14} which can be measured by doing simple arithmetic operations in this vector space.

In our novel word2vec extensions, we exploit the additive property of word2vec vectors in the context of fine-grained zero-shot learning.
A concrete example for this property is as follows~\cite{MYZ13}.
When we subtract the vector of \emph{man} from the vector of \emph{king} and add the vector of \emph{woman}, the resulting vector is closest to the vector of \emph{queen}.
Our fine-grained image classification task requires finding subtle differences between two words describing two different bird species.
In the following we assume that we have a list of attributes that name 
properties of different bird species.
Instead of asking for human judgement on how related a certain class, \ie \emph{black footed albatross}, and a certain attribute, \emph{curved beak}, we want to automatically determine this similarity using the vector differences of words in word2vec space.
We propose three {\it Noun-Attribute-Difference (NAD)} variants to capture 
relevant language information (sketched in Fig~\ref{fig:DeFrag}). 

The first version leads to a single language part, the second version consists of a constant number of parts per each class, and the third version leads to a variable number of language parts for each class. In the following formulations, we define a set of classes $C \in \{c_1,..,c_n\}$ with $n$ being the number of classes and a set of attributes $A \in \{a_1,..,a_m\}$ with $m$ being the total number of attributes. Moreover, $w2v(.)$ defines the vectorial representation of a word in the word2vec space. Accordingly, $wC(.)$ is the word2vec of a class and $wA(.)$ is the word2vec of an attribute.

\myparagraph{NAD1.} In this version, we aim to build a vector that represents the similarity of class words and attribute words in the semantic word2vec space. We define NAD1 as follows:
\begin{equation}
\text{NAD1}(c_i,j) = \|wC (c_i) - wA(a_j)\|, \forall a_j\in A 
\end{equation}
The NAD1 of a particular class is defined as the magnitude of the distance between the word2vec of a class and the word2vec of each attribute for all the classes.
As the number of attributes is fixed, there is a single NAD1 vector that is associated with each class. In other words, NAD1 corresponds to a single language part, \ie LP=1.

\myparagraph{NAD2.} As an alternative to using all attributes and all class names (NAD1), we aim to eliminate attributes that are not relevant for a particular class. Classically this is determined by human (expert) annotations which we want to avoid. 
Instead we argue that this human annotation effort can be eliminated by considering the similarity of class and attribute words in the word2vec space.

Based on the magnitude of the distance in the word2vec space, we define the set of attributes that are relevant for a class as follows. $B(c_i) = \{a_j | a_j \in A_{top-n}(c_i)\}$ where $A_{top-n}(c_i)$ is a set of attributes that are the $top-n$ nearest neighbors in word2vec space to class $c_i$. Accordingly, our second NAD version is formulated as follows:
\begin{equation}
\text{NAD2}(c_i) = \{wC(c_i) - wA(a_j) | a_j \in B(c_i)\} 
\end{equation}
The NAD2 leads to the same number of language parts for each class.
However, as for each class the most similar $top-n$ attributes are highly likely to be different, the set of attributes that are used in NAD2 is naturally not the same for each class.
We select $LP = \{5, 10, 25, 50, 75, 100\}$ and build six different sets of NAD2 representations.

\myparagraph{NAD3.} For the definition of the final alternative, we additionally assume that 
we know which attributes are present for which class even though we do not know how important an attribute is for any class. NAD3 is defined as follows:
\begin{equation}
\text{NAD3}(c_i) = \{wC(c_i) - wA(a_j) | a_j \in A(c_i)\}
\end{equation}
where $A(c_i)$ is the set of attributes associated to class $c_i$. In the experiments below $A(c_i)$ is obtained by thresholding the continuous attribute strengths which is known to introduce errors~\cite{RSS11}. 
It is important to note that only NAD3 requires set $A(c_i)$ and that the other two NAD variants only require the list of attributes that is relevant to {\it all} classes and 
use the similarities of attributes and classes in word2vec space to automatically generate language parts. 

\subsection{MBoW: Multiple Bag-of-Words} 
\label{subsec:multi-lang_w2v}

Similar to the NAD in Sec~\ref{subsec:multi-lang_w2v}, we build multiple language parts associated for each class as an extension to the BoW method.
We use wikipedia articles that corresponds to each class as our text corpus.
We build three different versions of multiple bag-of-words.

\myparagraph{MBoW1.} As a baseline, we extract a single BoW histogram from the entire wikipedia article of each class.
This leads to one language part per class.

\myparagraph{MBoW2.} Here, we divide the wikipedia articles of each class into a constant number of paragraphs.
This number is selected from the set $P = \{2,3,4,5\}$.
As the wikipedia article of each class has different length, the MBoW2 vectors that correspond to classes with shorter articles will get sparser with the increasing number of $P$.

\myparagraph{MBoW3.} As wikipedia articles have a structural organization of their own, in this version of multi-bag-of-words, we use this wikipedia structure.
We divide the articles into different subject-separated partitions.
As different articles have different number of sections, MBoW3 leads to a variable number of vectors for each class.

\section{Experiments}
\label{sec:experiments}

In our experimental evaluation we use the fine-grained Caltech UCSD Birds-2011 (CUB)~\cite{CaltechUCSDBirdsDataset} dataset that contains 200 classes of different North-American bird species populated with $\approx$60 images each. Each class is also annotated with 312 visual attributes. In the zero-shot setting, 150 classes are used for training and the other 50 classes for testing. For parameter validation, we also use a zero-shot setting within the 150 classes of the training set \ie we use 100 classes for training and the rest for validation.

We extract image features from the activations of the fully connected layers of a deep CNN. We re-size each image to 224$\times$224 and feed into the network which was pre-trained following the model architecture of the 16-layer VGG network~\cite{VL14}\footnote{We use the publicly-available MatConvNet library~\cite{VL14}}. As multiple visual parts, we use image features extracted from the annotated part locations of the images. For this, we crop the image on the overlapping bounding boxes with the size 50$\times$50~\footnote{We have empirically found that 50$\times$50 performs well for the task} that we draw around that particular part location (Sec.~\ref{sec:visual}), resize each bounding box to 224$\times$224 and follow the rest of the pipeline.

As supervised language parts (Sec.~\ref{sec:language}), we use human-annotated per-class attributes with continuous values that measure the strength of the attribute for each class. As unsupervised language parts we automatically extract word2vec~\cite{MSCCD13} from the entire 13.02.2014 wikipedia dump and Bag-of-Words from the wikipedia articles that correspond to our 200 object classes. For NAD, \ie our novel Noun-Attribute-Differences, we take the word2vec vectors of 200 classes and 312 attributes. For MBoW, we use the same vocabulary as before and extract BoW histograms from different parts of wikipedia articles.

\begin{table}[t]
 \begin{center}
  \begin{tabular}{l c| c || c | c|}
	\cline{3-5}
	 & & \textbf{Supervised} & \multicolumn{2}{c|}{\textbf{Unsupervised}}  \\
	\hline
	\multicolumn{1}{|l|}{\textbf{ Method}} & \textbf{VP}  & Attributes & word2vec & BoW \\
	\hline
	\multicolumn{1}{|l|}{Ours} & $1$& $43.3$ & $\mathbf{25.0}$ & $\mathbf{21.8}$ \\
	\hline
	\multicolumn{1}{|l|}{SJE~\cite{ARWLS15}}& $1$& $\mathbf{50.2}$ & $24.2$ & $20.0$ \\
	\hline
  \end{tabular}
 \end{center}
\caption{Comparison with the state-of-the-art using a single visual part (VP=1, we use only the whole image as a ``part'') and a single language part (LP=1) obtained using supervised attributes or unsupervised word2vec and BoW.}
\label{tab:methods}
\end{table}

\subsection{Effect of Learning Method} As a baseline for our evaluation, we employ Structured Joint Embedding (SJE)~\cite{ARWLS15} which learns a bilinear compatibility function between an image and its class embedding. SJE obtains the current state-of-the-art for zero-shot learning on CUB. We re-evaluate SJE using our 4K-dim VGG-CNN~\cite{VL14} as input embedding\footnote{Note that~\cite{ARWLS15} reports slightly better performance using GoogLeNet features instead of VGG as here.}. We use attributes as supervised output embeddings, word2vec and bag-of-words as unsupervised output embeddings. On the other hand, our joint part embedding framework learns two compatibility functions parameterized by $W^{language}$ and $W^{visual}$ with an integrated non-linearity computation. 
Tab.~\ref{tab:methods} compares SJE and our joint embedding 
using the standard average per-class Top-1 image classification accuracy on previously unseen classes. Using a single visual part per image, our joint embedding performs worse in the supervised setting (attributes) but slightly better than SJE in the unsupervised setting. Namely, joint part embeddings achieve $25.0\%$ for word2vec while SJE obtains $24.2\%$ and $21.8\%$ for Bag-of-Words whereas SJE obtains $20.0\%$ accuracy. Here, the language parts are extracted from wikipedia without using any human annotation. This result is important as we aim to increase the zero-shot learning performance on the CUB dataset for this unsupervised setting. The following section exploits our flexible framework
to incorporate both strong visual supervision as well as multiple language parts.

\begin{table}[t]
 \begin{center}
  \begin{tabular}{|c|c||c||c|c|}
	\hline
	 \textbf{Train} & \textbf{Test} & \textbf{Supervised} & \multicolumn{2}{c|}{\textbf{Unsupervised}} \\
	\cline{3-5}
	\textbf{VP} & \textbf{VP} & Attributes & word2vec & BoW \\
	\hline
	$1$ & $1$ & $43.3$ & $25.0$ & $21.8$ \\
	\hline
	 $19$ & $1$ & $47.0$ & $26.8$ & $22.6$ \\
	\hline
	 $19$ & $19$ & $\mathbf{56.5}$ & $\mathbf{32.1}$ & $\mathbf{26.0}$ \\
	\hline
  \end{tabular}
 \end{center}
\caption{Multiple visual parts (VP) for classification. VP are extracted from the annotations that are provided with the dataset. (Top-1 avg per-class top-1 acc on unseen classes.)}
\label{tab:multvisual}
\end{table}

\begin{table*}[t]
 \begin{center}
 \resizebox{\linewidth}{!}{
  \begin{tabular}{c c| c@{\hskip 2mm} c@{\hskip 2mm} c@{\hskip 1mm} | c@{\hskip 1mm}|| c@{\hskip 2mm} c@{\hskip 2mm} c@{\hskip 1mm} |c@{\hskip 1mm}| c@{\hskip 2mm} c@{\hskip 2mm} c@{\hskip 1mm} |c@{\hskip 2mm}|}
	\cline{3-14}
	 & & \multicolumn{4}{c||}{\textbf{Supervised}} & \multicolumn{8}{c|}{\textbf{Unsupervised}}  \\
	\hline
	\multicolumn{1}{|c|}{ \textbf{Train}} & \textbf{Test} & \multicolumn{4}{c||}{Attributes} & \multicolumn{4}{c|}{word2vec} & \multicolumn{4}{c|}{BoW} \\
	 \cline{3-14}
	 \multicolumn{1}{|c|}{ \textbf{VP}} & \textbf{VP} &R@1&R@5 &R@10 &mAUC &R@1 &R@5 &R@10 &mAUC &R@1 &R@5 &R@10 &mAUC\\
	\hline
	\multicolumn{1}{|c|}{$1$} & $1$ & $47.0$ & $91.7$& $95.9$& $36.5$& $37.5$& $55.6$& $73.2$& $22.8$& $33.2$& $49.8$& $61.0$ &$16.2$ \\
	\hline
	\multicolumn{1}{|c|}{$19$} & $1$ & $65.7$ & $87.7$& $91.8$ & $38.7$& $40.6$& $59.0$& $67.3$ & $24.5$& $30.8$& $46.6$& $57.0$ & $17.3$ \\
	\hline
	\multicolumn{1}{|c|}{$19$} & $19$ & $ 61.6$ & $93.9$ &$100.0$ & $\mathbf{46.6}$& $43.1$& $69.5$& $71.5$ &$\mathbf{30.7}$ & $30.6$& $48.6$& $50.7$& $\mathbf{22.0}$ \\
	\hline
  \end{tabular}
   }
 \end{center}
\caption{Multiple visual parts (VP) for retrieval. VP are extracted from the annotations that are provided with the dataset. We measure recall at 1,5,10 (R@1,5,10) and mean AUC on unseen classes.}
\label{tab:retrieval}
\end{table*}

\subsection{Strong Supervision by Part Annotations}
Apart from using a non-linear embedding objective, our joint part embedding benefits from using multiple visual or language parts. We extract 19 parts from each image that correspond to the whole image, head, body and full bounding box~\cite{ZDGD14}, bounding boxes drawn around 15 part locations whose annotations are available within the dataset. We evaluate the effect of parts in the following way: (1) training and testing with a single part, (2) training with multiple parts and testing with a single part, and (3) training and testing with multiple parts.

\myparagraph{Zero-Shot Image Classification.} For zero-shot image classification, we calculate the mean per-class Top-1 accuracy obtained on unseen classes. In other words, we consider the prediction as positive only if the predicted class label matches the correct class label for that image. We average the predictions on a per-class basis. The results are presented in Table~\ref{tab:multvisual}. For attributes, using multiple visual parts at training time already improves the accuracy from $43.3\%$ to $47.0\%$, improving the state-of-the-art. On the other hand, using multiple visual parts also at test time achieves $56.5\%$ accuracy, further improving the supervised state-of-the-art on this dataset. For Bag-of-Words, using multiple visual parts improves the accuracy $26.0\%$. For word2vec, multiple visual parts achieves an impressive $32.1\%$ accuracy which becomes the new state-of-the-art obtained without using human supervision on the language side. These results support our intuition that using strong supervision of semantic visual parts leads to more discriminative image representations and thus is helpful for zero-shot fine-grained image classification.

\myparagraph{Zero-Shot Image Retrieval.} For zero-shot image retrieval, we use two popular evaluation methods: the average recall at position 1, 5 and 10 (R@1,R@5 and R@10) on the ranked list of labels predicted for each image and the mean area under the Precision-Recall curve (mAUC). We present our results in Tab~\ref{tab:retrieval}. The state-of-the-art~\cite{BSFS15} retrieval accuracy reported on unseen classes without human supervision on the CUB dataset is $13.0\%$ mAUC. Using VP=1 both BoW (mAUC of $16.2\%$) and word2vec ($22.8\%$) outperform the state-of-the-art. Using $VP=19$ 
further improves performance for BoW ($22\%$) and word2vec ($30.7\%$). Using supervised text annotation, \ie attributes, the unseen class mAUC increases to $46.6\%$. These results indicate that strong visual supervision helps both image retrieval and classification in a zero-shot learning setting.

\begin{figure}[t]
\centering
   \includegraphics[width=0.8\linewidth, trim=30 30 40 70,clip,angle=-90,origin=c]{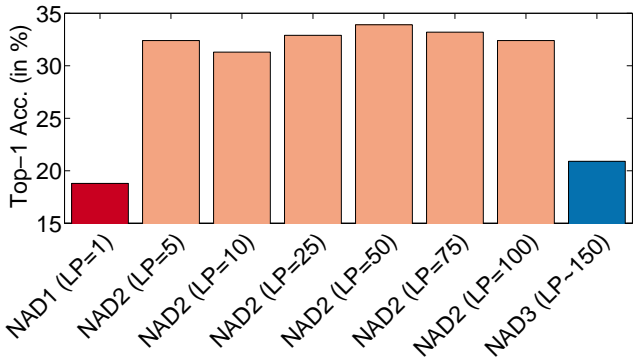}
      \vspace{-35mm}
   \caption{Effect of multiple language parts. \emph{NAD} uses noun-attribute distance as a measure of similarity.}
   \label{fig:w2v}
\end{figure}

\subsection{Using Multiple Language Parts} We now explore the effects of using multiple language parts and to associate them to multiple visual parts extracted using strong supervision. 

\myparagraph{NAD.} We evaluate the three proposed noun-attribute distance (NAD) variants (Sec~\ref{subsec:multi-lang_w2v}) as language parts. NAD1 measures noun-attribute distances between all classes and all attributes and results in a single language part (LP=1). For NAD2, the noun-attribute distances are computed between all classes and top 5-100 most discriminative attributes. Thus, it corresponds to LP=5-100. NAD3, measuring noun-attribute distances between all classes and all the relevant attributes for that class. Therefore, NAD3 uses different number ($\approx150$) of language parts for each class.

We present our results with NAD on Fig~\ref{fig:w2v}. 
NAD1 and NAD3 both do not obtain impressive results. In the case of NAD1 this can be explained with the fact that it only contains a single language part. For NAD3 we suspect that 
this is due to the fact that there is a large imbalance in the number of descriptive attributes for each class.
NAD2 on the other hand obtains promising results. In fact using 50 language parts (NAD2 LP=50) 
obtains $33.9\%$ (see also Tab.~\ref{tab:final}) that improves over the previous unsupervised state-of-the-art using word2vec alone.

\begin{figure}[t]
\centering
   \includegraphics[width=0.78\linewidth, trim=30 30 40 70,clip,angle=-90,origin=c]{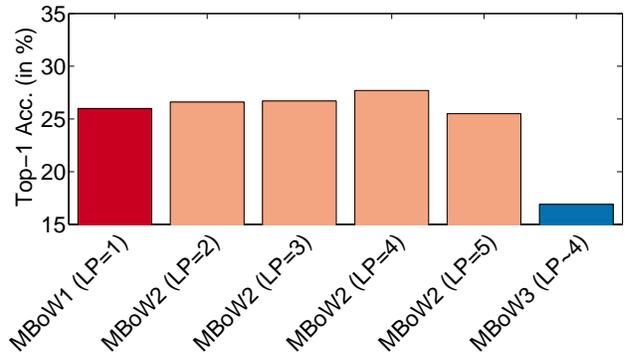}
   \vspace{-32mm}
   \caption{Effect of multiple language parts. \emph{MBoW} uses multiple parts from wikipedia articles.}
   \label{fig:bow}
\end{figure}

\myparagraph{MBoW.} As an alternative multiple language parts setting to NAD, we use MBoW (Sec~\ref{subsec:multi-lang_w2v}) also extracted three different ways. For MBoW1, we construct the BoW using the entire wikipedia article for a bird which results in a single language part (LP=1). For MBoW2, same number of multiple language parts (LP=2-5) are extracted by partitioning the wikipedia articles into 2,3,4 or 5 parts. For MBoW3, we extract variable number of language parts ($\approx 4$) based on wikipedia's own content-grouped paragraphs.

We present our results on Fig~\ref{fig:bow}. 
MBoW1 as well as the different versions of MBoW2 obtain reasonable performance even though staying below the best
performance achieved by NAD2. The results of MBoW3 are not that impressive. As this representation 
is based on the wikipedia article structure itself and as mentioned before the length of the paragraphs 
is variable and thus the histogram based representation might be not reliable enough.

\def\arraystretch{1.3}
\begin{table}[t]
 \begin{center}
  \begin{tabular}{l l|c||c|c|}
  	\cline{2-5}
  	& \multicolumn{4}{|c|}{\textbf{Visual Parts (VP)}} \\
	\cline{2-5}
	&\multicolumn{1}{|l|}{\textbf{Train}$\;\rightarrow$} & $1$ & $19$ & $19$ \\
	\cline{2-5}
	&\multicolumn{1}{|l|}{\textbf{Test}$\;\downarrow\rightarrow$} &  $1$ & $1$ & $19$ \\
	\hline
	\multicolumn{1}{|c}{\textbf{Language Parts}}& \multicolumn{1}{|c|}{$1$} & $25.0$ & $26.8$ & $32.1$ \\
	\cline{2-5}
	\multicolumn{1}{|c}{\textbf{(LP)}}& \multicolumn{1}{|c|}{$50$} & $23.6$ & $30.5$ & $\mathbf{33.9}$  \\
	\hline
  \end{tabular}
 \end{center}
\caption{Summary of our results with single or multiple visual and language parts. We improve over the state-of-the-art with unsupervised embeddings significantly.}
\label{tab:final}
\end{table}
\def\arraystretch{1}

\myparagraph{Summary of results.} We investigate the effects of using a single visual part for training + test and using multiple visual parts either only on training or both on training + test. The results are summarized in Tab~\ref{tab:final}. In our framework, we can use a single visual part in combination with either a single language part or multiple language parts. The former configuration leads to $25.0\%$ accuracy whereas the latter obtains a lower accuracy of $23.6\%$. This indicates that multiple language parts help only if they are supported with multiple visual parts.

Another interesting configuration is using multiple visual parts during training and, at test time, evaluating the multiple language parts in these cases with or without strong visual annotation. In the former case, again, there are two configurations,\ie with a single language part and with multiple language parts. With a single language part we obtain $26.8\%$ accuracy which is already higher than $25.0\%$ with a single visual part at training time. On the other hand, with multiple language parts we achieve an impressive $30.5\%$ accuracy. This shows that multiple language parts indeed help even if they are supported by strong visual supervision only at training time. If we use multiple visual parts also at test time, we further improve our results to $33.9\%$, establishing a new state-of-the-art when using unsupervised text embeddings.

\newcommand{\vv}{\checkmark}
\begin{table}[t]
 \begin{center}
  \begin{tabular}{|c|c | c | c|c|c|}
	\hline
	 & \multicolumn{3}{c|}{\textbf{Unsupervised}}& \multicolumn{2}{c|}{\textbf{Test}}\\
	\hline
	 \textbf{LP} &  W2V & BoW & NAD2 & VP=$1$ & VP=$19$ \\
	\hline
	\multirow{3}{*}{$\mathbf{1}$} & $\vv$ &       &       & $26.8$ & $32.1$\\
	 &       & $\vv$ &       & $22.6$ & $26.0$\\
	 & $\vv$ & $\vv$ &       & $33.2$ & $\mathbf{34.7}$\\
	\hline
	\multirow{3}{*}{$\mathbf{50}$} &      &       & $\vv$ & $30.5$ & $33.9$\\
	 & $\vv$ &       & $\vv$ & $31.0$ & $32.1$\\
	 &       & $\vv$ & $\vv$ & $30.0$ & $\mathbf{34.3}$\\
	\hline
  \end{tabular}
\end{center}
\caption{Combining different number of language parts (LP). \emph{word2vec:} class-based vectors extracted from wikipedia. \emph{BoW:} histogram of word occurrences per wikipedia article of a class. \emph{NAD2:} Using noun-attribute distance as a measure of similarity between classes. We use multiple visual parts at training and either single or multiple visual parts at test time.}
\label{tab:ensemble_all}
\end{table}

\subsection{Multi-Cue Language Embeddings}

We finally explore combinations of different language parts in our joint part embeddings framework. For single language part setting (LP=1), we combine word2vec and BoW. For training, we use multiple visual parts (VP=19), whereas for testing we either use multiple visual parts (VP=19) or single visual part (VP=1). The results are presented in Table~\ref{tab:ensemble_all}. Combining word2vec with BoW using VP=1 for testing leads to $33.2\%$ accuracy improving both word2vec ($32.1\%$) and BoW ($26.0\%$). Additionally, the same combination with VP=19 for testing leads to $34.7\%$ accuracy which again improves word2vec ($32.1\%$) and BoW ($26.0\%$) on the same setting. These results are consistent and encouraging because they provide a large improvement over the state-of-the-art ($24.2\%$, Table~\ref{tab:methods}) and reduces the gap between the state-of-the-art obtained through human annotation ($50.2\%$, Table~\ref{tab:methods}).

For the setting with multiple language parts, we use the best performing NAD2 with VP=50. This method measures the similarity of word2vec vectors between class and attribute names with the most relevant (top50) attributes to each class. Using a single visual part for testing leads to $30.5\%$ accuracy whereas using multiple visual parts obtains $33.9\%$ accuracy. Compared to the single-part word2vec ($32.2\%$) and BoW ($26.0\%$), this is a significant improvement which indicates that combining multiple language parts also help. Moreover, word2vec contains latent relationships between class and attribute names which are released when these nouns are considered relative to each other.

Finally, the last row of Table~\ref{tab:ensemble_all} shows that the combination of NAD2 ($33.9\%$) and BoW ($26.0\%$) leads to $34.3\%$ accuracy which is again higher than NAD2 and BoW alone. This indicates that our approach can exploit the complementarity of the NAD2 and BoW representations.

\section{Conclusion}
\label{sec:conclusion}
For the challenging problems of zero-shot fine-grained classification and retrieval, we have presented a formulation that allows to integrate diverse class descriptions and detailed part annotations and consequently improves significantly on the state-of-the-art on both tasks in a range of experimental conditions. In particular, we have demonstrated how to compensate for the loss of accuracy by using weaker auxiliary information with detailed visual part level annotations. Our approach facilitates a joint embedding of multiple language parts and visual information into a joint space. With strong visual supervision and human-annotated attributes we improve the state-of-the-art on the CUB dataset to $56.5\%$ (from $50.2\%$) in the supervised setting. In addition, we show how to use multiple language sources and extract diverse auxiliary information from unlabeled text corpora, \ie word2vec and BoW. We build multiple parts on the language side, \ie NAD and MBoW and thereby improve the state-of-the-art also in the unsupervised setting to $33.9\%$ (from $24.2\%$). Finally, we combine different unsupervised text embeddings and further improve the results for the unsupervised setting to $34.7\%$. 

As a conclusion, we propose several extensions for fine-grained zero-shot learning. 
First, using multiple visual parts when available, \ie training or test time, rather than using a single visual part 
leads to a significant boost in performance. 
Second, these multiple visual parts can be supported with multiple language parts for further improvements. 
Third, word2vec space indeed contains some latent information and distance between class and attribute names can eliminate the costly human annotation of class-attribute associations. 
Following these practices, we improve the fine-grained zero-shot state-of-the-art on CUB for both supervised and unsupervised text embeddings.

\small
\bibliographystyle{ieee}
\bibliography{myrefs,mateuszref,mario}

\end{document}